\documentclass{article}

\usepackage[round]{natbib}

\usepackage{microtype}
\usepackage{graphicx}
\usepackage{subfigure}
\usepackage{booktabs} 
\usepackage[utf8]{inputenc} 
\usepackage[T1]{fontenc}   
\usepackage{url}        
\usepackage{booktabs}   
\usepackage{amsfonts}      
\usepackage{nicefrac}     
\usepackage{lscape}
\usepackage{comment}
\usepackage{svg}
\usepackage{comment}
\usepackage{amsmath}
\usepackage{amssymb}
\usepackage{csquotes}
\usepackage{xcolor}
\usepackage{amsthm}
\usepackage{pascaldef}
\usepackage{siunitx}
\usepackage{mathtools}
\usepackage{array}
\newcolumntype{L}{>{$}l<{$}}
\usepackage{tikz}
\usetikzlibrary{positioning, fit, arrows.meta, shapes}
\usepackage{algpseudocode}

\newcommand{\R}{\mathbb{R}}
\DeclareMathOperator{\KL}{KL}
\newcommand{\KLe}[2][]{\operatorname{\widehat{KL}}^{#1}_{#2}}
\newcommand{\He}[2][]{\operatorname{\widehat{H}}^{#1}_{#2}}
\newcommand{\Data}{\mathcal{D}}
\newcommand{\noise}{\boldsymbol{\epsilon}}
\newcommand{\Like}{\mathcal{L}}
\newcommand{\V}{\mathcal{V}}
\newcommand{\batch}{\mathcal{B}}
\newcommand{\std}[1]{\scriptsize$\pm\, \num[round-mode=places,round-precision=1]{#1}$}
\newcommand{\txtparagraph}[1]{\textbf{#1}\ }

\theoremstyle{plain} 
\newtheorem{theorem}{Theorem}[section] 
\newtheorem{proposition}[theorem]{Proposition}

\theoremstyle{definition}
	
\theoremstyle{remark}

\usepackage{hyperref}

\usepackage[accepted]{icml2021}

\icmltitlerunning{Out-of-distribution detection for regression tasks: parameter versus predictor entropy}

\begin{document}

\twocolumn[
\icmltitle{Out-of-distribution detection for regression tasks: \\
parameter versus predictor entropy}




\begin{icmlauthorlist}
\icmlauthor{Yann Pequignot}{to}
\icmlauthor{Mathieu Alain}{to}
\icmlauthor{Patrick Dallaire}{to}
\icmlauthor{Alireza Yeganehparast}{to}
\icmlauthor{Pascal Germain}{to}
\icmlauthor{François Laviolette}{to}
\end{icmlauthorlist}

\icmlaffiliation{to}{Université Laval, Québec, Canada}

\icmlcorrespondingauthor{}{yann.pequignot@iid.ulaval.ca}

\icmlkeywords{Machine Learning, ICML}

\vskip 0.3in
]



\printAffiliationsAndNotice{}  

\begin{abstract}
It is crucial to detect when an instance lies downright too far from the training samples for the machine learning model to be trusted, a challenge known as out-of-distribution (OOD) detection. For neural networks, one approach to this task consists of learning a diversity of predictors that all can explain the training data. This information can be used to estimate the epistemic uncertainty at a given newly observed instance in terms of a measure of the disagreement of the predictions. Evaluation and certification of the ability of a method to detect OOD require specifying instances which are likely to occur in deployment yet on which no prediction is available. Focusing on regression tasks, we choose a simple yet insightful model for this OOD distribution and conduct an empirical evaluation of the ability of various methods to discriminate OOD samples from the data. Moreover, we exhibit evidence that a diversity of parameters may fail to translate to a diversity of predictors. Based on the choice of an OOD distribution, we propose a new way of estimating the entropy of a distribution on predictors based on nearest neighbors in function space. This leads to a variational objective which, combined with the family of distributions given by a generative neural network, systematically produces a diversity of predictors that provides a robust way to detect OOD samples. 
\end{abstract}
\section{Introduction}
As more and more industries are adopting artificial intelligence (AI) technologies, the requirements surrounding the development and deployment of AI-based systems are also evolving. In the medical sector and in aerospace for instance, several exciting applications have been developed, but will not make it to the real world until guarantees are obtained that critical mistakes can be avoided or at least safely controlled~\citep{bhattacharyya2015certification, Begoli2019}. Quantification of uncertainty is one important mechanism to support guarantees, therefore contributing to the safety of a system~\citep{shafaei2018uncertainty}, as long as it is accurate and exhaustive. Of particular interest is the ability of a system to detect instances which are \enquote{too far} from the data it learned from for the prediction to be trusted, a task known as out-of-distribution (OOD) detection. 

When learning a neural network, many different parameters (\emph{i.e.} weights and biases) may be able to explain the data with similar accuracy while disagreeing strongly on other instances. This under-specification phenomenon is leveraged by several methods which seek to learn not only a single parameter but a set or a distribution of them, an information which can be used to estimate the epistemic uncertainty at any newly given instance in terms of the dispersion of the corresponding predictions. This uncertainty prediction can in turn be used to discriminate OOD instances from in-distribution ones. However, learning a faithful diversity of predictors is a challenging task.

The conceptually simplest method for learning a set of predictors relies on the randomness of the learning process and samples several parameters by performing stochastic gradient descent from a random initialization \citep{lakshminarayanan2017simple}. In a different direction, MC dropout \citep{gal2016dropout} consists of learning a single parameter encoding a stochastic predictor that relies on a random zeroing out of entries after every hidden layer. In the Bayesian approach~\citep{Mackay1992, Neal1996}, the choice of an a \textit{prior} distribution on all predictors determines through Bayes' rule a distribution of predictors for every given training set: the \textit{posterior} distribution.  

Markov Chain Monte Carlo (MCMC) methods such as Hamiltonian Monte Carlo (HMC) methods~\citep{Betancourt2017} form a golden standard for inferring the posterior distribution. However, in most cases for neural networks, this inference is difficult and computationally very expensive to carry out due to the non-linearity and the typically large dimension of the parameter space. Meanwhile, variational methods have gradually emerged as a time-efficient solution for obtaining an approximate posterior distribution via an optimization procedure~\citep{Bishop2006, Wainwright2008, Graves2011}. They address the problem of intractability by finding a good approximate distribution among a given family of distributions, the simplest method based on this idea being 
Mean Field Variational Inference
(MFVI) 
which is known as \emph{Bayes by Backprop}  \citep{Blundell2015} when turned into a training algorithm for neural networks.

In this paper we focus on uncertainty prediction for regression tasks and evaluate several methods on OOD detection. We choose an OOD distribution in order to evaluate qualitatively the predicted uncertainty, compute ROC curves and AUC estimator. As it turns out that MFVI performs poorly, we investigate two directions to explain this failure. First, in order to increase the capacity of the variational family, we consider implicit distributions given by generative networks. Second, we question the objective of MFVI which indirectly aims at a diversity of parameters. Using our chosen OOD distribution, we suggest an alternative objective which directly aims at a diversity of predictors. 

\section{Regression, uncertainty and OOD detection}

We consider a regression task given by a set $\Data$ of independent observation pairs $(X,y)$ with input $X\in \R^D$ and target $y\in\R$. All methods considered in this paper produce models that predict a set $Y(X)$ of values for any given input $X$. This set $Y(X)$ allows for a prediction $U(X)$ of the epistemic uncertainty of the model based on a measure of the dispersion of $Y(X)$. To detect OOD instances, one ultimately chooses a threshold $s$ to classify an instance $X$ as OOD if $U(X)>s$ or as in-distribution otherwise. 

Given an in-distribution sample and an OOD sample, one can evaluate the OOD detection capability of a model by considering the associated Receiver Operator Curve (ROC) and its Area Under the Curve (AUC), which estimates the probability that the model will predict a larger uncertainty on a randomly chosen OOD sample than on a randomly chosen in-distribution sample. 
However, the choice of the OOD distribution is specific to each application case and it is a complicate and delicate one. The study presented in this paper is based on a generic proxy for the OOD distribution: 
Given in-distribution sample as input data, we propose to use a uniform distribution $\nu$ on a rectangle based on the input data to serve as a proxy for the OOD. More specifically, OOD samples were produced by sampling each feature uniformly from $[x_\text{min}, x_\text{max}]$ where $x_\text{min}$ and $x_\text{max}$ are respectively the minimum and maximum value taken by this feature in the dataset. 


Concerning the epistemic uncertainty $U(X)$ at some input $X$, we make a Gaussian assumption about the distribution of $Y(X)$ to estimate its differential entropy. In other words, we use $U(X)=\frac{1}{2}\ln(2\pi e \sigma^2)$ where $\sigma^2$ is the unbiased estimate of the variance of $Y(X)$. Note that this choice is equivalent to using directly the variance $\sigma^2$ as far as the AUC is concerned.

\section{Bayesian Inference}
 
To find a model that predicts for every $X\in \R^D$ (of interest) a distribution over the possible targets $y\in \R$ representing best our knowledge given our postulates and requirements, the Bayesian regression approach proposes to first assume that the data are generated according to:
\begin{equation} \label{eq:BayesInferencePostulate}
y=f(X)+\epsilon\,,
\end{equation}
where $f:\R^D\to \R$ is an unknown deterministic function and $\epsilon$ is a random variable with density $\mathcal{N}(0,\sigma_l)$. Moreover, in order to reflect that in most applications inputs do not all have the same importance, we consider that the input $X$ in \eqref{eq:BayesInferencePostulate} is a random variable that follows either the unknown true distribution of input data or a context driven distribution that allows for an explicit focus on OOD inputs.

Postulate \eqref{eq:BayesInferencePostulate} induces the likelihood of any function $g:\R^D \to \R$ as $\Like(g|\Data)=\textstyle\prod_{(X,y)\in \Data} \Like(g|X,y)\,,$ with $ \Like(g|X,y)=p(y|X,g)=p_\epsilon(y-g(X))\,,$ where $p_\epsilon$ is the density function of the random variable $\epsilon$.  One then further postulates or chooses a distribution $P$ on all functions/predictors $g:\R^D\to \R$, the prior. Altogether, this allows one to define another distribution on predictors, the posterior distribution $P(g|\Data)$, whose density is given by Bayes' rule:
$
dP(g|\Data)=\frac{\Like(g|\Data)}{p(\Data)}dP(g)\, ,
$
where the \emph{marginal likelihood} (also called \emph{evidence} and \emph{integrated likelihood}) of the data is the usually intractable normalizing constant:
$p(\Data)=\int \Like(g|\Data) dP(g)\,.$
The prediction of uncertainty about the target at input $X$ is obtained by integration with respect to the predictive posterior  
as:
\[
p(y|X,\Data)=\frac{1}{p(\Data)}\int p(y|X,g)\, dP(g|\Data)\,.
\]
As the evidence is in most cases intractable, we are left with the task of approximating the posterior distribution $Q$ so as to compute the predictive uncertainty with the best accuracy. In this direction, we record the following simple and well-known fact \citep[\eg,][\textsection 21.2]{murphybook} which gives a characterization of the posterior distribution in variational terms as the unique solution to an optimization problem based on the Kullback-Leibler (KL) divergence (for completeness, the proof is given in the supplementary material).
\begin{proposition}[Variational characterization of the Bayesian posterior] 
\label{BayesPosteriorVariationalDef}
For every likelihood function $\Like(g|\Data)$ and every prior distribution $P$, the Bayesian posterior distribution is the unique distribution $Q$ that minimizes the variational objective:
\begin{equation*}
\textstyle\V(Q)=-\Esp_{g\sim Q}\bigl[\ln \Like(g|\Data)\bigr]+ \KL(Q,P)\,.
\end{equation*} 
\end{proposition}
\txtparagraph{Tractability, Predictor Representation through Neural Networks.}
In practice, we do not have direct access to \emph{all} the possible predictive functions $g:\R^D\to\R$. We have only access to some of them through a chosen representation, that is, a parametric model family. Given the success of neural networks in machine learning, they have been extensively used to model predictors in regression tasks. A network architecture comes with a specific number of parameters, say $d$, consisting of \emph{weights} and \emph{biases}. This gives rise to the parameter space $\mathbb{R}^d$; henceforth, a vector $\theta\in \mathbb{R}^d$ represents an ordered list of weights and biases for the network and hence encodes a unique predictor $f_\theta:\mathbb{R}^D\to \mathbb{R}$. 

By identifying a parameter $\theta$ with the corresponding function $f_\theta$, it becomes possible to implement the Bayesian approach as follows. We choose a prior distribution $P$ on parameters with tractable density function $p(\theta)$, for instance a fully factorized Gaussian distribution. We define the likelihood of a parameter $\theta$ as the likelihood of $f_\theta$, \emph{i.e.} $p(y|X,\theta)=p(y|X,f_\theta)$ and $\Like(\theta|\Data)=\Like(f_\theta|\Data)$. Then we define the density function of the posterior distribution on parameters up to a scaling factor:
$
p(\theta|\Data) \propto \Like(\theta|\Data)p(\theta)
$.
Since the right-hand term is tractable and (almost everywhere) differentiable, it can be used to approximate the posterior distribution using MCMC methods such as Hamiltonian Monte Carlo (HMC) methods~\citep{Betancourt2017}. These methods yield samples $(\theta_j)_{j<N}$ of parameters which can be used to estimate the predictive uncertainty:
\begin{equation*}
p(y|X,\Data)=\!\!\int\!\! p(y|X,\theta) p(\theta|\Data)  d\theta \approx \!\!\frac{1}{N} \sum_{j<N}\!\! p(y|X,\theta_j). 
\end{equation*}
Note, however, that by identifying a parameter with the function it represents, we have traded the infinite dimensional geometry of functions for the \textit{ad-hoc} $d$-dimensional Euclidean geometry of the parameters. In particular, if $f_{\theta_0}$ and $f_{\theta_1}$ are two functions yielding very similar predictions on some data, what can we say about the Euclidean distance between $\theta_0$ and $\theta_1$? 
In particular, we have traded the infinite dimensional geometry of functions for the \textit{ad-hoc} $d$-dimensional Euclidean geometry of the parameters. In particular, if $f_{\theta_0}$ and $f_{\theta_1}$ are two functions yielding very similar predictions on some data, what can we say about the Euclidean distance between $\theta_0$ and $\theta_1$? 

\txtparagraph{Parametric Variational Inference.} Prop.~\ref{BayesPosteriorVariationalDef} opens an avenue for approximating the Bayesian posterior. Given the prior distribution~$P$ and the likelihood function $\Like(\theta|\Data)$, we choose a parametric family $\mathcal{Q}=\{Q_\lambda\mid \lambda \in \R^l\}$ of distributions on the space $\R^d$ of parameters and search for the distribution $Q\in\mathcal{Q}$ that minimizes the opposite of the \emph{evidence lower bound} (ELBO):
\begin{equation}
\V(Q)=-\Esp_{\theta\sim Q}\bigl[\ln \Like(\theta|\Data)\bigr]+ \KL(Q,P)\,.
\label{Eq:ParametricVI}
\end{equation}
By substituting this variational objective for an estimator that is differentiable w.r.t. the parameters of the variational distribution $Q$, this minimization can be achieved using stochastic gradient descent (SGD). The need for such an estimator of the variational objective usually restricts the choice of the variational family $\mathcal{Q}$. 
The first term of Eq.~\eqref{Eq:ParametricVI}, the expected log likelihood of the variational distribution~$Q$ given the data, rewards distributions that concentrate on functions that predict the data accurately. It is easily estimated by a Monte Carlo (MC) estimate as long as we can sample from $Q$. Moreover this term does not suffer the fact that we identify $f_\theta$ with $\theta$, since the likelihood of $\theta$ is defined as the likelihood of $f_\theta$. The second term of Eq.~\eqref{Eq:ParametricVI}, the KL divergence from our variational distribution $Q$ to our prior $P$, rewards distributions that are close to $P$.
Moreover it rewards variational distributions whose differential entropy $H(Q)$ is large since 
$
\KL(Q,P)=-H(Q)- \Esp_{\theta \sim Q} \ln p(\theta)\,,
$
where $\ln p(\theta)$ denotes the log density of the prior.
The latter term is harder to estimate and the computation of an MC estimate requires to have access to the log density of both the prior $P$ and the variational distribution $Q$ in addition to being able to sample from $Q$.

Variational families for which the density is available in closed form are called \emph{explicit families}. As these methods tend to restrict to families based on products of independent Gaussian distributions~\citep{ranganath2014black, Blundell2015, Saul1996, Barber1999} for computational reasons, they often lead to an underestimation of uncertainty~\citep{turner2011}. Several means have been deployed to improve the capacity of explicit families such as structured or mixture families \citep{Saul1996,Bishop2006}, but recently various methods have been proposed to implement variational inference for \emph{implicit families}. These families trade off a density available in closed form for higher capacity and more flexibility. Among these, we have families for which the density function exists and can be efficiently computed with \emph{invertible networks} and \emph{normalizing flows}~\citep{kingma2016,dinh2016density,papamakarios2017masked}. Finally, we have the implicit families for which the density function cannot be computed or may not even exist. The optimization in that case relies on other ideas such as Spectral Stein gradient, kernel density estimation \citep{li2017gradient}, or even adversarial density ratio estimation \citep{mescheder2017adversarial}. In the next section, we present two ways to optimize an arbitrary implicit family, the first one for distributions over parameter space, the second one for distributions over predictor space.

\section{Proposed Method}

We consider implicit families of distributions given by generative networks and consider two alternative objectives: one in parameter space, and one in predictor space.


Since 
the probability density function of our variational distribution is intractable both in parameter space and predictor space,\footnote{Note that even for Mean-Field families (MFVI) whose density function is tractable in parameter space, the density is no longer tractable in predictor space (see FuNN-MFVI).
} we choose to rely on a nearest neighbor estimate of the KL term in both cases.  

\begin{figure*}[ht!]
    \centering
    \includegraphics[width=1\textwidth]{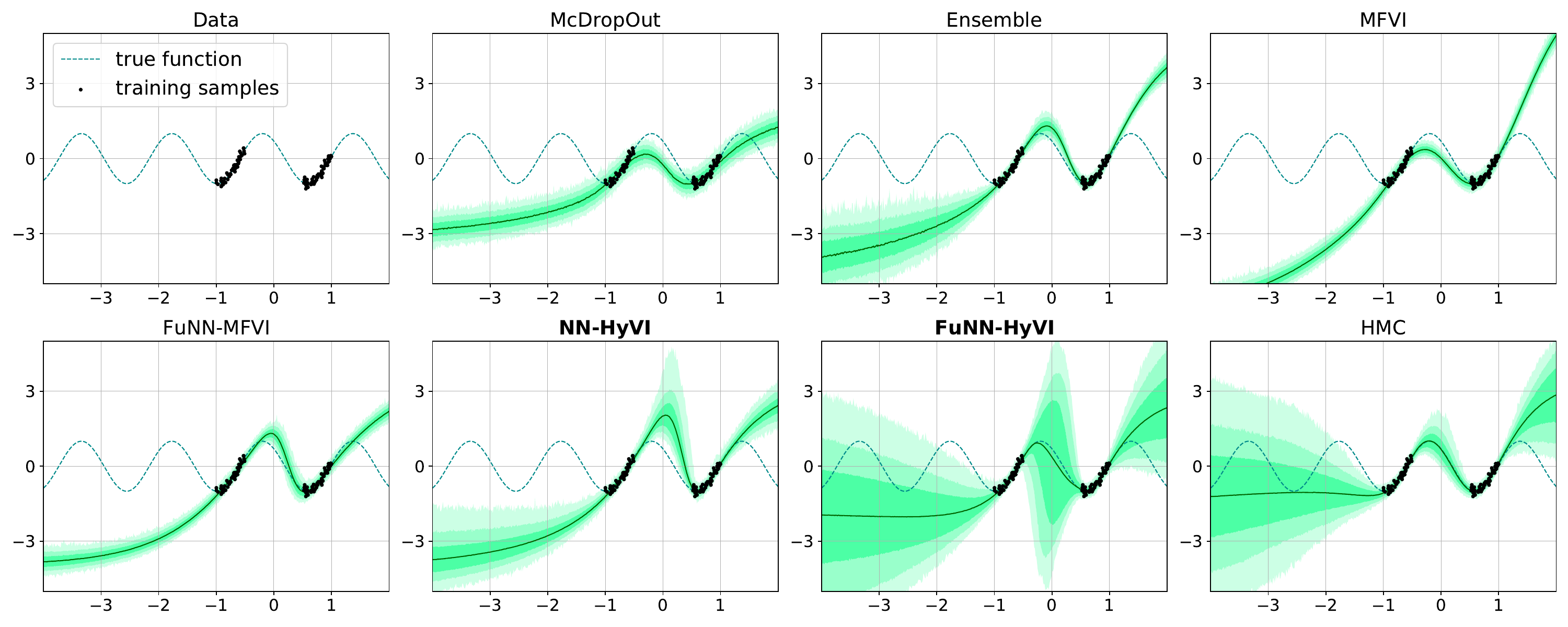}
    \caption{Synthetic dataset: Comparison of various models for prediction of uncertainty. The solid line represents the mean of the predictions and each shade of green represents one standard deviation.}
    \label{fig:toy_example_ood}
\end{figure*}

\subsection{Nearest Neighbor Estimation of KL Divergence and Entropy}

Drawing inspiration from nearest neighbor (NN) estimates of the KL divergence and the differential entropy in parameter space, we present a simple approach to estimating those quantities for distributions on predictors.

\textbf{The Parameter Space.}
Given $d$-dimensional independent 
samples 
$\tilde{Q}=\{q_i\}_{i<N}$
and 
$\tilde{P}=\{p_j\}_{j<M}$ 
from distributions $Q$ and $P$ respectively, the k-NN estimation of the KL divergence from $Q$ to $P$ is given by \citep[e.g.][]{Wang2009DivergenceEF}:
\begin{equation}
\KLe{k}(\tilde{Q},\tilde{P}) = 
\ln \tfrac{M}{N-1} +\tfrac{d}{N}\textstyle\sum_{i<N}\ln \tfrac{s_k(q_i)}{r_k(q_i)}\,,
\label{KLestimator}
\end{equation}
where $r_k(q_i)$ and $s_k(q_i)$ denote the Euclidean distance from $q_i$ to its $k$-nearest neighbor in $\tilde{Q}\setminus \{q_i\}$ and in $\tilde{P}$ respectively. While low values of $k$ are in principle favored to reduce bias, increasing the value of $k$ allows us to avoid underflow imprecision when samples are very close to each other.  

Importantly, the KL divergence from $Q$ to $P$ can be decomposed as $\KL(Q,P){=\,-}H(Q)- \Esp_Q[ \ln P]$ where $H(Q)$ denotes the differential entropy.
The following provides an estimator of the differential entropy of $Q$ based on samples 
$\tilde{Q}=\{q_i\}_{i<N}$ 
along the same line \citep{Singh2003}:
\begin{equation}\label{HNNestimator}
\He{k}(\tilde{Q})= C_{d,k,N}+ \tfrac{d}{N}\textstyle\sum_{i<N} \ln r_k(q_i)\,,
\end{equation}
where $C_{d,k,N}{=}\ln(N)-\psi(k)+\ln\bigl(\nicefrac{\pi^{\frac{d}{2}}}{\Gamma(\frac{d}{2}+1)}\bigr)$ is a constant, $\psi(x)=\nicefrac{\Gamma(x)'}{\Gamma(x)}$ is the digamma function.

\textbf{The Predictor Space.}
Let $\Pi\in \R^d$ and $\Theta\in \R^d$ be random variables following the prior distribution $P$ and the variational distribution $Q$ respectively. Instead of relying on $\KL(Q,P)$, we would like to consider the \enquote{KL divergence} from the distribution of $f_\Pi$ to that of $f_\Theta$. Since the KL estimator \eqref{KLestimator} relies solely on relative distances of the samples, it opens an avenue for formalizing this idea by substituting the Euclidean distance between parameters with a distance on functions which reflects the geometry of our predictors. What distance do we want to consider on the space of predictors? As we shall see, the Euclidean distance between two parameters $\theta$ and $\theta'$ may differ drastically from the distance between the corresponding functions $f_\theta$ and $f_{\theta'}$. For instance, because of symmetries in our network-based model, we may have $f_\theta=f_{\theta'}$ while $\Vert\theta-\theta'\Vert$ is large. It may also happen that $f_\theta(X)=f_{\theta'}(X)$ for
most inputs $X$ in the domain of interest while the parameters $\theta$ and $\theta'$ are far apart in the Euclidean~space. 

We therefore suggest choosing a distribution $\nu$ on the input space $\R^D$ that represents the relative importance we give to all possible inputs. This choice depends on the specific task and is closely related to the choice of the samples used for the evaluation of OOD detection. Therefore in our experiments, we take $\nu$ to be the distribution chosen for the OOD evaluation, namely the uniform distribution whose support is a bounded hyperrectangle constructed from the training data.

Choosing a distribution $\nu$ on the inputs allows us to define the \emph{predictor space} as the Hilbert space $L_2(\nu)$ with the norm distance:
\[
\Vert f -g \Vert_{L_{2}}=\bigl(\textstyle\int | f(X)-g(X)|^2\,d\nu(X)\bigr)^\frac{1}{2}
\,.
\]
Note that every sample $\mathcal{X}=(X_0,\ldots X_{T-1})\sim \nu^T$ yields a MC estimate of the $L_2$ norm
$\Vert f -g \Vert_{L_{2}} \approx \big(\frac{1}{T} \textstyle\sum_{t<T} | f(X_t)-g(X_t)|^2\big)^\frac{1}{2}= \frac{1}{\sqrt{T}}\Vert f^{\mathcal{X}}-g^{\mathcal{X}}\Vert_2$ 
in terms of the Euclidean distance $\Vert \cdot \Vert_2$ between the vectors $f^{\mathcal{X}}=(f(X_0),\ldots,f(X_{T-1}))\in \R^T$ and $g^{\mathcal{X}}=(g(X_0),\ldots,g(X_{T-1}))\in \R^T$ obtained by evaluation. Hence each input sample $\mathcal{X}$ yields an evaluation map $f \mapsto f^\mathcal{X}$ which provides an approximation from the infinite dimensional Hilbert space $L_2(\nu)$ in the Euclidean space $\R^T$ up to a scaling factor of $T^{-\nicefrac{1}{2}}$.

Therefore given independent samples\footnote{In our experiments, the sample of functions $(f_i)$ is obtained from a sample of parameters $(\theta_i)$ by letting $f_i$ be the function encoded by $\theta_i$ via our network, \emph{i.e.} $f_i=f_{\theta_i}$.} $\tilde{F}=(f_i)_{i<N}$ and $\tilde{G}=(g_j)_{j<M}$ in $L_2(\nu)$ from random variables $F$ and $G$ respectively, our proposed KL divergence estimate is
\begin{equation} \label{FunKLestimator}
\KLe[\nu]{k} (\tilde{F},\tilde{G})
=\textstyle\Esp_{\mathcal{X}\sim \nu^T} \KLe{k}(\tilde{F}^\mathcal{X},\tilde{G}^\mathcal{X})\,,
\end{equation} where the estimator from Eq.~\eqref{KLestimator} is applied to the $T$-dimensional samples $\tilde{F}^\mathcal{X}=\{f_i^\mathcal{X}\}$ and $\tilde{G}^\mathcal{X}=\{g_i^\mathcal{X}\}$. Note that no adjustment to Eq.~\eqref{KLestimator} is necessary because it only involves ratios of distances, and the scaling factor $T^{-\nicefrac{1}{2}}$ cancels out.

In the same fashion, we propose to estimate the differential entropy of the random variable $F$ from samples 
$\tilde{F}=\{f_i\}_{i<N}$ 
in $L_2(\nu)$ via
\begin{equation} \label{FunHNNestimator}
\He[\nu]{k} =\textstyle\Esp_{\mathcal{X}\sim \nu^T} \He{k}(\tilde{F}^\mathcal{X})-\frac{1}{2}\ln T \,,
\end{equation}
where the constant term comes from our scaling of distances by the factor $T^{-\nicefrac{1}{2}}$.

\subsection{Implicit Variational Inference}
A generative network is simply a neural network architecture encoding a parametric family of functions $h_{\lambda}:\mathbb{R}^l \to \mathbb{R}^d\,,$ where the parameter $\lambda$ lists the \emph{weights and biases} of the neural network and each $h_\lambda$ transforms a Gaussian random noise $\noise {\sim} \mathcal{N}(0, \mathbf{I}_l)$ into another random variable in $\R^d$. Any parametric family $(h_\lambda)_{\lambda\in \Lambda}$ of networks therefore gives rise to the family of random variables of the form $\boldsymbol{\theta}=h_\lambda(\noise)$. In our context, this serves as our variational family and the outputs $\theta$ of the generative network are parameters for the predictor network $f_\theta$. We therefore refer to the generative network as hypernet and talk about Hypernet Variational Inference (HyVI) to highlight the difference with the family of multivariate Gaussian with diagonal covariance matrix and Mean Field Variational Inference (MFVI, aka Bayes By Backprop).

\txtparagraph{In the Parameter Space: NN-HyVI.}
The first of our methods (Nearest Neighbor -- Hypernet Variational Inference) performs inference directly in the parameter space. Given a predictor network architecture $y=f_\theta(X)$, a (possibly implicit) prior $P$ on parameters $\theta$ and a likelihood function $\Like(f|X,y)$, we update the parameters of the variational distribution $q_\lambda$ using SGD so as to minimize on each mini-batch $\mathcal{B}$ of training data $\Data$ the objective (see 
Suppl. Mat.).
\begin{equation}\label{ObjectiveNN}
    \V(q_\lambda)=
    \tfrac{|\batch|}{|\Data|}\Esp_{\substack{\theta\sim q_\lambda \\ \pi\sim P}}\!\!\KLe{1}(\theta,\pi)-\sum_{\mathclap{(X,y)\in\batch}}\Esp_{{\theta\sim q_\lambda}}\!\!\!\ln \Like(f_\theta|X,y)  \,.
\end{equation}
\txtparagraph{In the Predictor Space: FuNN-HyVI.}
Our second method (Functional Nearest Neighbor -- Hypernet Variational Inference) performs inference directly in the predictor space, and uses the parameter space only as a convenient way to represent the predictor space. It also relies on a parametric model of predictors $y=f_\theta(X)$, but it further requires a distribution $\nu$ on inputs from which we can sample from. Given also a (possibly implicit) prior $P$ on predictors and a likelihood function $\Like(f|X,y)$, we update the parameters of a variational distribution $q_\lambda$ using SGD so as to minimize  on mini-batch $\mathcal{B}$ of training data $\Data$ the following objective (see 
Suppl. Mat.):
\begin{equation}\label{ObjectiveFuNN}
    \V(q_\lambda)=\tfrac{|\batch|}{|\Data|}\Esp_{\substack{\theta\sim q_\lambda \\ g\sim P}}\!\!\!\KLe[\nu]{1}(f_\theta,g) {-}\sum_{\mathclap{(X,y)\in\batch}}\Esp_{\theta\sim q_\lambda}\!\!\! \ln \Like(f_\theta|X,y).
\end{equation}
Since we only need to sample predictors from the prior, our method allows for the use of a prior given by a Gaussian process, for example. However in our experiments, we used a prior $P$ on parameters and sampled predictors according to $g=f_\pi$ with $\pi\sim P$.

\subsection{Related Work}
\txtparagraph{Implicit Weight Uncertainty in Neural Networks.} \cite{pawlowski2017implicit} also used an NN estimate of the KL divergence on parameters. However, it appears that they treat each parameter independently and use an average of the KL approximations over each scalar parameter. This approach lacks a theoretical justification and the authors' preference is only based on limited empirical evidence of its efficiency.
In contrast, we address the important dependency between the parameters by using the NN estimator of the KL on parameter vectors.

\txtparagraph{FBNN.} A closely related work by \citet{grosse2019} also provides a method for variational inference in predictor space. They give an interesting theoretical justification for an expression of the KL divergence from one stochastic process to another as the supremum over all finite sets of inputs of the KL divergence between corresponding marginal distributions. However, they end up using an approximation which consists of averaging over input samples the KL divergence of the corresponding marginal distributions. While this approximation is not theoretically justified in their paper, it turns out to be similar to our approximation of the KL for distributions of functions \eqref{FunKLestimator} which naturally arise from a novel motivation for an explicit choice of the input distribution for the MC approximation of the $L_2$ distance on predictors. Moreover, they use inputs from the training data in addition to OOD inputs to marginalize predictors and a Stein Spectral Gradient estimate of the gradient of the KL on marginal distributions. Two crucial differences with our FuNN-HyVI are 1) they use a variational family of multivariate Gaussian distribution with diagonal covariance matrix (as in our FuNN-MFVI which performs poorly) where we use a generative network and 2) they regularize the KL term to increase its importance in the ELBO on mini-batches where we maintain the expected gradient of the ELBO. Finally, their method also differs in that they use data-dependent Gaussian process priors where we simply use a Gaussian prior on parameters.  

\section{Empirical Study}

Our experiments aim primarily at evaluating the quality of the uncertainty predicted with our methods. In accordance with the literature, we report the root-mean-square error (RMSE) and log Posterior Predictive (LPP) on held out test data (see Suppl. Mat.), but we also consider other metrics to evaluate the uncertainty of the model as a whole and as well as its ability to detect OOD inputs.

\txtparagraph{Entropy of \enquote{Posterior distributions}} For Bayesian methods, estimated the differential entropy of the posterior distributions based on $1K$ samples using~\eqref{HNNestimator}, and in predictor space using an MC estimate of \eqref{FunHNNestimator} based on $100$ samples from $\nu^{200}$ (cf. Tab.~\ref{Table:EntropyOfPosteriors}). For the sake of comparison, we made similar estimations for Ensemble by treating the set of parameters as a sample of a posterior distribution. Note that for MC dropout we do not have access to a set of parameters; however we can still evaluate the diversity of the model in predictor space similarly by considering for each sample from $\nu^{200}$ a set of $1K$ predictions.

\label{sec:empirical_study}

\subsection{Experimental Setup}

We experiment on a synthetic dataset and 8 UCI regression datasets (5 small ones, 3 large ones). 

\label{Subsec:setups}
\txtparagraph{Synthetic Data.}
We generated a training set consisting of $120$ pairs $(X,y)$ with inputs $X$ sampled uniformly from $[-1.,-0.5] \cup [0.5,1.]$ using the function $
y {=} \cos(4(X{+}0.2)){+}\epsilon\,,$ with $\epsilon \sim \mathcal{N}(0., 0.1)$. The OOD distribution $\nu$ is uniform on $[-4.,2.]$.

\txtparagraph{UCI Datasets.}
From the UCI machine learning repository \citep{uci}, we use five \emph{small datasets}: Boston Housing ($D=13$, $N=506$), Concrete Compressive Strength ($8$,$1030$), Energy ($8$, $768$), Wine Quality Red ($11$, $1599$), Yacht Hydrodynamics ($6$, $308$), and three \emph{large datasets}: Power Plant Combined Cycle ($4$, $9568$), Condition Based Maintenance of Naval Propulsion Plants Compressor ($16$, $11034$) and Protein Data ($9$, $45730$). We  split each dataset in training (90\%) and test (10\%). 

\txtparagraph{Predictor Network.} For the synthetic dataset, we use a single layer network with $50$ hidden units and Tanh activations. For the UCI datasets, we use a single layer with ReLU activations and $50$ hidden units for the small ones, $100$ for the large ones.

\txtparagraph{Prior.} In all our experiments, we use the same Gaussian prior on the parameters of the predictor network with zero mean and covariance matrix $0.5\mathbf{I}$. FuNN-HyVI allows us to specify a prior in the form of a Gaussian process, but we have not observed any significant improvement as far as OOD detection is concerned.   

\renewcommand{\std}[1]{\scriptsize$\pm\, \num[round-mode=places,round-precision=2]{#1}$}

\subsection{Experiments}

\txtparagraph{Synthetic example.} We compare various methods to predict uncertainty on our synthetic dataset, using when possible $\sigma_l=0.1$ (the noise chosen for the data generation) for the likelihood.

\txtparagraph{UCI datasets.} For each UCI dataset, we choose a specific random train/test split and use for the scale $\sigma_l$ of the Gaussian likelihood a fixed value inspired by a grid search in \citet{gal2016dropout} (see Suppl. Mat.). Note that Ensemble does not use theses noises since it simply optimizes the square loss.  

We used HMC to estimate the posterior distribution on each small dataset and every other method is run three times with random initialization on all datasets. 

\subsection{Implementation details}

\txtparagraph{Generative Network Training (HyVI).} For both NN-HyVI and FuNN-HyVI, we use a generative network with ReLU activations transforming $5$-dimensional Gaussian samples $\mathcal{N}(0,\mathbf{I}_5)$ via a first layer with $20$ hidden units and a second layer with $40$ hidden units (the number of output units equals the number of parameters of the predictor network). 
This hypernet is trained using Adam optimizer, with a learning rate in $[0.005, 0.0001)$ decreasing by a ratio of $0.7$ when no progress is made in the variational objective after a patience of $30$ epochs ($10$ for the large datasets). We use mini-batches of size $50$, except for the large datasets, where we use a size of $500$.

\txtparagraph{NN-HyVI.}
We use $100$ samples of the variational distribution for evaluating the expected log likelihood. For estimating the KL with $\KLe{1}$, we use $500$ samples from each of the variational distribution and the prior. 

\txtparagraph{FuNN-HyVI.}
We similarly sample $100$ times from the variational distribution for evaluating the expected log likelihood. For approximating the KL in predictor space using $\KLe[\nu]{1}$, we sample $500$ times from each of variational distribution and the prior and evaluate these parameters at a single sample from $\nu^T$. On the synthetic examples, we use $T=50$, whereas on the UCI datasets we use $T=200$.

Other methods include HMC \citep{Betancourt2017}, MC dropout \citep{gal2016dropout}, Ensemble \citep{lakshminarayanan2017simple}, MFVI \citep{Blundell2015} (details in Suppl. Mat.) and the following variant of MFVI.

\txtparagraph{FuNN-MFVI.} 
The Gaussian posterior parameters are trained using the same objective as in FuNN-HyVI~\eqref{ObjectiveFuNN} and same mini-batch size, optimizer and learning rate scheme.

\begin{table}[t]
\setlength{\tabcolsep}{4pt}
\center \scriptsize
\begin{tabular}{lccccccc}
\toprule
& &  MC & & & FuNN- &NN-& FuNN-\\
{} &     HMC &  dropout &  \!\!\!Ensemble\!\!\! &    MFVI &  MFVI &  HyVI &  HyVI \\
\midrule
boston   &  \bfseries 1.0000 &      0.9476 &   \bfseries  1.0000 &  0.9670 &     0.9964 &   0.9953 &     \bfseries 1.0000 \\
concrete &  0.9957 &      0.9254 &    \bfseries 0.9998 &  0.9343 &     0.9891 &   0.9927 &     0.9977 \\
energy   &  0.9998 &      0.7483 &    \bfseries 1.0000 &  0.4104 &     0.9665 &   0.9999 &  \bfseries    1.0000 \\
wine     & \bfseries  1.0000 &      0.9869 &    0.9997 &  0.9964 &     0.9957 &   0.9996 &     0.9996 \\
yacht    &  0.9890 &      0.4836 &    \bfseries 0.9958 &  0.5438 &     0.9443 &   0.8888 &     0.9826 \\
\midrule
navalC      & NA&    \bfseries     1.0000 &  \bfseries   1.0000 &  0.8669 &     0.9578 &  \bfseries  1.0000 &    \bfseries  1.0000 \\
powerplant & NA&     0.9144 &    0.9335 &  0.8635 &     0.9335 &   0.9572 &     \bfseries 0.9626 \\
protein    & NA&       0.9999 &    0.9999 &  0.9988 &     0.9966 &   \bfseries 1.0000 &    \bfseries  1.0000 \\
\bottomrule
\end{tabular}
 \caption{Average AUC score for each method the UCI datasets. }
    \label{tab:AUC}
\end{table}

\begin{table}[t]\center
	\scriptsize
	\setlength{\tabcolsep}{4pt}

	\begin{tabular}{lccccccc}
	\toprule
& &  MC & & & FuNN- &NN-& FuNN-\\
{} &     HMC &  dropout &  \!\!\!Ensemble\!\!\! &    MFVI &  MFVI &  HyVI &  HyVI \\
	\midrule
	{} &\multicolumn{7}{c}{\bfseries Parameter Space} \\
    \midrule
boston   &  970 &    NA &  332 &   567 &  -1117 & \bfseries 1555 &     350 \\
concrete &  635 &    NA &  424 &   353 &    -697 & \bfseries  1078 &      431 \\
energy   &  628 &    NA &  193 &   348 &   -465 & \bfseries 1098 &     334 \\
wine     &  615 &    NA &  812 &  247 &  -1035 &\bfseries 1455 &  -1596 \\
yacht    &  502 &    NA &  152 &   263 &    -589 & \bfseries   772 &     145 \\
    \midrule
navalC     &  NA &    NA &  636 &  -400 &  -2553 & \bfseries 1726 &  -4777 \\
powerplant &   NA &    NA &   286 &     366 &    -832 & \bfseries 1066 &      53 \\
protein    &   NA &    NA &  1205 &     788 &   -2189 &\bfseries  2314 &  -1222 \\
\midrule
    {} & \multicolumn{7}{c}{ \bfseries Predictor Space (w.r.t to $\nu$)}\\
    \midrule
  boston   &   -5 &   -491 &   -171 &   -317 &   122 &   -398 & \bfseries 294 \\
concrete &  -107 &  -492 &  -211 &   -316 &    76 &   -423 & \bfseries  273 \\
energy   &  -224 &  -506 &  -346 &   -587 &  -10 &  -301 & \bfseries 160 \\
wine     &   -25 &  -234 &      7 &  -422 &  128 &  -285 & \bfseries 254 \\
yacht    &  -244 &   -844 &   -574 &   -492 &  -83 &   -667 &\bfseries  113 \\
\midrule
navalC     &   NA &   -167 &    -62 &  -719 &  \bfseries 136 &  -511 &   12 \\
powerplant &   NA &  -607 &  -598 &   -571 &   -74 &  -760 &  \bfseries 169 \\
protein    &   NA &   -18 &     93 &  -367 &  155 &    -71 & \bfseries 372 \\
	\bottomrule
	\end{tabular}
	\caption{Average entropy of posterior distributions.} 
	\label{Table:EntropyOfPosteriors}
\end{table}

\subsection{Results}
In this section, we raise empirical evidence that carrying out the inference in predictor space, such as done by FuNN-HyVI, indeed yields a higher epistemic uncertainty of the whole model, a property which translates into robustness for OOD detection. In particular, this suggests that variational inference in predictor space could lead to higher quality uncertainty than the computationally demanding golden standard provided by computationally expensive MCMC methods, such as HMC, which are restricted to the parameter space of the network.

\txtparagraph{Synthetic Data.}
Fig.~\ref{fig:toy_example_ood} shows the uncertainty predicted by the various methods on our synthetic dataset. We find that MC dropout, MFVI, FuNN-MFVI all fail to estimate the uncertainty. Ensemble struggles to predict higher uncertainty on OOD, especially in the region between the two patches. NN-HyVI predicts higher uncertainty on OOD but still provide an underestimation on the boundaries of the interval ($X\!=\!-4$ and $X\!=\!2$) in comparison with HMC and FuNN-HyVI. Finally, we find that FuNN-HyVI (runtime: 12 sec.) provides an uncertainty prediction that is quite close to HMC (runtime: 15 hours) while being even less confident on OOD, especially in the region between the two patches.

\txtparagraph{AUCs.}
Average AUC on three independently trained models for each method are displayed in Table~\ref{tab:AUC}. It was computed using the whole input data as in-distribution and $10K$ samples from the distribution $\nu$ for the OOD. On the small datasets, HMC, Ensemble and FuNN-HyVI perform extremely well, while MC dropout and MFVI achieve the worst scores. While scores for MC dropout improve on large datasets, MFVI still performs unsatisfactorily. On the large datasets, FuNN-HyVI always displays the best performance, with NN-HyVI close by. We note that these two methods of ours outperform Ensemble by a significant margin on the powerplant dataset. While FuNN-MFVI performs relatively well, it is always outperformed by FuNN-HyVI. Similarly, NN-HyVI always achieves a higher than MFVI. This provides evidence that the flexibility of our generative network is an important asset--even by trading the full support of the Gaussian for a low (at most $5$-dimensional) submanifold support. Moreover, we observe that the AUC of NN-HyVI is always lower than that of FuNN-HyVI, which supports the idea that carrying inference in predictor space yields a more reliable uncertainty for OOD detection.

\begin{figure*}[t]
\centering
\includegraphics[width=0.65\textwidth]{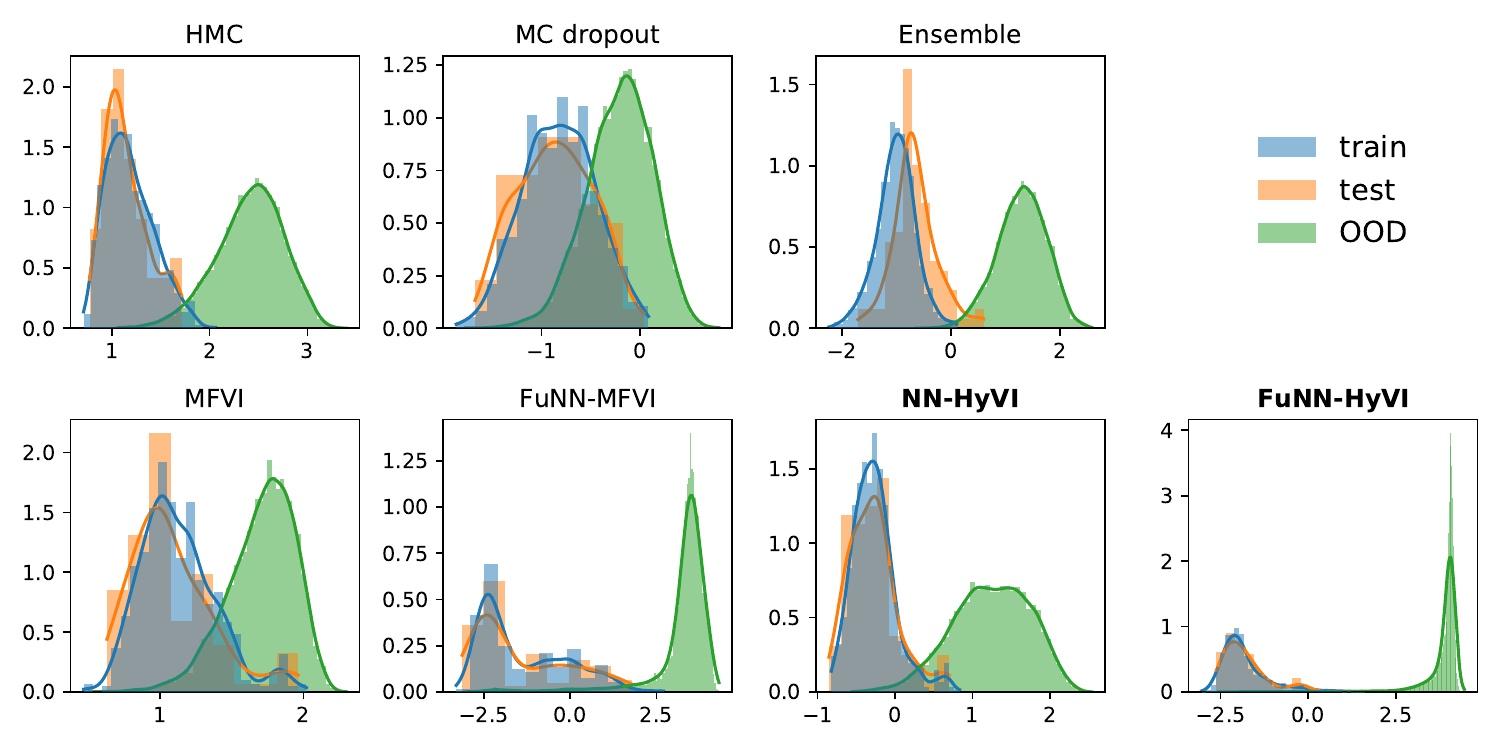}
\includegraphics[width=0.33\textwidth]{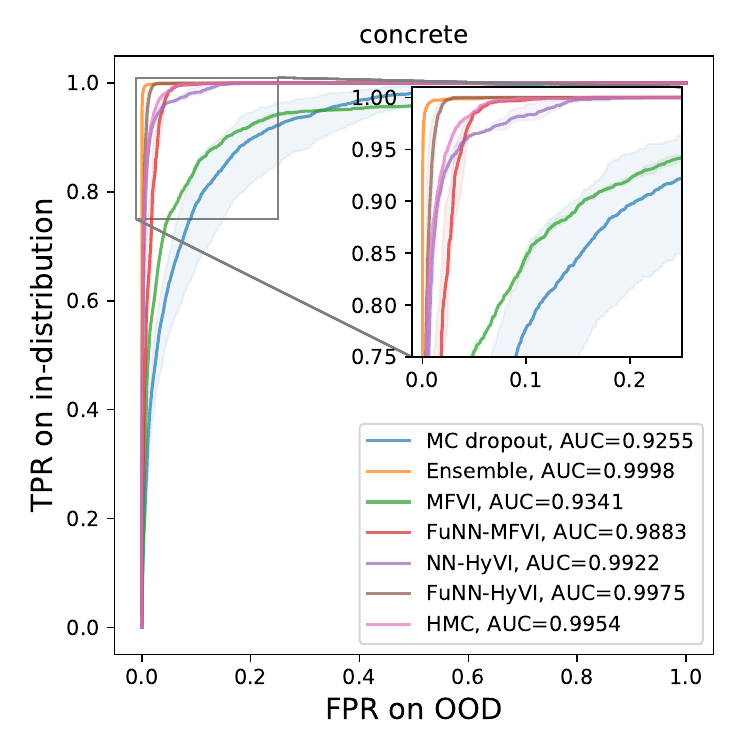}
 \caption{Distributions of the epistemic uncertainty on train, test and OOD inputs on the Concrete UCI dataset and the corresponding ROC curves (the line represents the mean and the shade area represents min and max values over the three models).  Similar results for all datasets are in supplementary material.}
    \label{fig:EntropyConcrete}
\end{figure*}

\txtparagraph{Entropy of posterior distributions.}
We report in Table~\ref{Table:EntropyOfPosteriors} the average entropy of posterior distributions on the three independently trained models. When considered as distribution on parameters, on all datasets NN-HyVI achieves a larger entropy than every other method, while FuNN-HyVI always admits a low entropy. On the contrary, in predictor space, FuNN-HyVI achieves the largest entropy (except on the navalC dataset). In particular, when comparing NN-HyVI and FuNN-HyVI which both optimize the same hypernet family, we see that the inference in predictor space makes sure that FuNN-HyVI always achieves a larger diversity of predictors. As for NN-HyVI and MFVI which both optimize the entropy in parameter space, MFVI always captures less parameter diversity. While this corroborates the advantage of our generative network over the family of Gaussian with diagonal covariance matrix, we see that a larger parameter diversity does not necessarily mean a larger predictor diversity.  More generally, these results provide evidence that there is no direct relation between the entropy in parameter space and the entropy in predictor space. In this respect, note that while HMC and NN-HyVI both carry out inference in parameter space, HMC performs better since it achieves a larger entropy in predictor space with a smaller entropy in parameter space. However, FuNN-HyVI has the best performance in this~respect. We consider that this also supports our claim that the inference in the parameter space is made (artificially) hard by the \textit{ad hoc} representation of predictors induced by the choice of the network architecture.
Moreover, we suggest that the entropy of the posterior in predictor space is a good measure of the epistemic uncertainty of the overall variational model with respect to the chosen OOD distribution. 

\txtparagraph{Robustness of uncertainty prediction.}
The distributions of epistemic uncertainty predicted on train, test and OOD inputs are displayed in Fig.~\ref{fig:EntropyConcrete} for each model, together with the corresponding ROC curves. We note that the AUC is only one possible summary statistic of the ROC curve and that, furthermore, the ROC curve does not captures fully the discrepancies between the distributions of epistemic uncertainty. In particular, while Ensemble achieves an AUC similar to that of FuNN-HyVI on all datasets (except powerplant), the corresponding distributions of epistemic uncertainty are always quite different. In fact, the uncertainty produced by FuNN-HyVI puts the OOD samples much further apart from the in-distribution samples than the one produced by Ensemble. Bearing in mind that one needs to choose a threshold in order to actually implement OOD detection, this property of FuNN-HyVI makes it more robust than Ensemble.  




\section{Conclusion}
Every method for predicting uncertainty with neural networks rely on a~specific way to collect a diversity of predictors. MCMC methods use a random walk in parameter space, Ensemble methods use the randomness inherent to initialization and training, MC dropout optimizes a stochastic network, variational inference methods rely on maximizing the differential~entropy. While most methods seek diversity in parameter space, we found empirical evidence that a diversity of parameters does not mean a diversity of predictors. Given the parametric class of predictors represented by a neural network and a dataset, the ultimate uncertainty is delicate to define and it certainly depends on the specific use case as well as relevant \emph{a priori} knowledge. In order to evaluate the uncertainty of our model, we choose an OOD distribution and design a detection task. We find that HMC and Ensemble perform well, while MFVI and MC dropout perform poorly. We present evidence that the poor performance of MFVI can be greatly improved by either using a more flexible distribution in the form of a generative network or by considering an alternative objective which maximizes the entropy in predictor space. This leads us to our method FuNN-HyVI which, in addition to performing very well in terms of AUC score, predicts a more robust uncertainty for OOD detection.


 \subsubsection*{Acknowledgments}
This work is supported by the DEEL Project CRDPJ 537462-18 funded by the National Science and Engineering Research Council of Canada (NSERC) and the Consortium for Research and Innovation in Aerospace in Québec (CRIAQ), together with its industrial partners Thales Canada inc, Bell Textron Canada Limited, CAE inc and Bombardier inc.\footnote{\url{https://deel.quebec}}

\bibliography{example_paper}
\bibliographystyle{icml2021}

\appendix

\onecolumn

\section{Proof of Proposition~\ref{BayesPosteriorVariationalDef} (Variational definition of the Bayesian posterior distribution)}

{\it
For every likelihood function $\Like(g|X,y)$ and every prior distribution $P$, the Bayesian posterior distribution is the unique distribution that minimizes the variational objective
\begin{equation}
\V(Q)=-\Esp_{g\sim Q}\bigl[\ln \Like(g|\Data)\bigr]+ \KL(Q,P)\,.
 \label{eq:BayesPosteriorVariationalDef}
\end{equation} 
}

\begin{proof}
Recall that for every distribution $Q$ and $Q'$ we have $\KL(Q,Q')\geq 0$ with equality iff $Q{=}Q'$. Let $Q$ denote the Bayesian posterior distribution and note that for every distribution $Q'$, we have
\[
0\leq \KL(Q', Q)=\int \ln\left(\frac{dQ'}{dQ}\right)\,dQ'=\KL(Q',P)- \int \ln \Like(g|\Data)\, dQ'(g)+ \ln p(\Data)\,,
\]
with equality iff $Q=Q'$. Hence for every distribution $Q'$ we have
\[
-\ln p(\Data)\leq \KL(Q',P)- \int \ln \Like(g|\Data)\, dQ'(g)\,,
\]
with equality iff $Q=Q'$. Since $p(\Data)$ does not depend on $Q'$, it follows the Bayesian posterior is the unique distribution that minimizes \eqref{eq:BayesPosteriorVariationalDef}, as desired.
 \end{proof}

\section{Pseudo code}
We include pseudo-code for NN-HyVI and FuNN-HyVI as Algorithm~\ref{alg:GenVI} and \ref{alg:GenVI-Pred}.
\begin{algorithm} 
  \begin{algorithmic}[1]
    \Require{data $\Data$, likelihood $\Like(f|X,y)$, prior distribution $P$}
    \Require{parametric predictive model $y=f_\theta(X)$}
    \Require{generative network $\theta=h_\lambda(\epsilon)$ with noise $\epsilon\sim \mathcal{N}$ and initial parameters $\lambda$}
   
    \While{\text{$\lambda$ has not converged}}
    
        \State $\epsilon_i \sim \mathcal{N}$, $i<N$ \Comment{Sample noise}
        
        \State $\theta_i= h_\lambda(\epsilon_i)$ \Comment{Compute parameters}
        
        \State $\batch\subseteq \Data$ \Comment{Get training mini-batch}

        \State $\text{LL}=\frac{1}{N}\sum_{i<N} \sum_{(X,y)\in \batch}\ln \Like(\theta_i|X,y)$ \Comment{Expected log-likelihood}
        
        \State $\pi_j \sim P$, $j<N$ \Comment{Sample from prior}

        \State $\text{KL}=\KLe{1}((\theta_i)_{i<N},(\pi_j)_{j<N})$
        \Comment{KL approximation} 
        
        \State $\V(\lambda)=\frac{|\batch|}{|\Data|} \text{KL}-\text{LL}$ \Comment{Compute ELBO for mini-batch}
        
        \State $\lambda \gets \text{Optimizer}(\lambda, \nabla_\lambda \V)$   
        
    \EndWhile
  \end{algorithmic}
    \caption{NN-HyVI
    \label{alg:GenVI}}
\end{algorithm}

\begin{algorithm}
  \begin{algorithmic}[1]
    \Require{data $\Data$, likelihood $p(y|X,f)$, prior distribution $P$}
    \Require{parametric predictive model $y=f_\theta(X)$}
    \Require{generative network $\theta=h_\lambda(\epsilon)$ with noise $\epsilon\sim \mathcal{N}$ and initial parameters $\lambda$}
    \Require{input distribution $\nu$}

    \While{\text{$\lambda$ has not converged}}
    
        \State $\epsilon_i \sim \mathcal{N}$, $i<N$ \Comment{Sample noise}
        
        \State $\theta_i= h_\lambda(\epsilon_i)$, $i<N$ \Comment{Compute parameters}
        
       \State $\batch\subseteq \Data$ \Comment{Get training mini-batch}

       \State $\text{LL}=\frac{1}{N}\sum_{i<N} \sum_{(X,y)\in \batch}\ln \Like(\theta_i|X,y)$ \Comment{Expected log-likelihood}
        
        \State $\mathcal{X}=(X_t)_{t<T} \sim \nu^T$ \Comment{Sample inputs}   
        \State $g_j \sim P$, $j<N$ \Comment{Sample from prior}   
        
        \State $f_{\theta_i}^\mathcal{X}=(f_{\theta_i}(X_t))_{t<T},\, g_j^\mathcal{X}=(g_j(X_t))_{t<T}, \, j<N$
        \Comment{Evaluate predictors at these inputs}
        \State $\text{KL}=\KLe{1}((f_{\theta_i}^\mathcal{X})_{i<N},(g_j^\mathcal{X})_{j<N})$
        \Comment{KL approximation} 
        
        \State $\V(\lambda)=\frac{|\batch|}{|\Data|} \text{KL}-\text{LL}$
        \Comment{Compute ELBO for mini-batch}
        
        \State $\lambda \gets \text{Optimizer}(\lambda, \nabla_\lambda \V)$   
        \Comment{Update parameters} 
        
    \EndWhile
  \end{algorithmic}
    \caption{FuNN-HyVI
    \label{alg:GenVI-Pred}}
\end{algorithm}

\section{Implementation details}

\paragraph{HMC.}
We consider the posterior distribution given by HMC \citep{Betancourt2017} as a reference.
 In each case, the thinning was adjusted to retain a maximum of 10,000 samples. We used dual averaging to tune the step size and checked for convergence and efficiency using $R$-hat and effective sample size (ESS) for tails and bulk \citep{vehtari2020} .
 
We use the code of the \texttt{minimc} library \citep{minimc} with the parameters presented in Tab.~\ref{parameters_HMC}.

\begin{table*}[ht!]
\small
\begin{center}
\begin{tabular}{l|ccc}
Dataset&number of iterations& burning & leapfrog steps\\
 \hline
 
Wave OOS Example & 120000 & 20000 & 200 \\  
Boston Housing & 170000 & 40000 & 100  \\
Concrete Compressive Strength & 170000 & 40000 & 100\\
Energy & 140000 & 4000 & 100\\
Wine Quality Red & 90000 & 20000 & 200\\
Yacht Hydrodynamics & 140000 & 40000 & 100 \\

\end{tabular}
\end{center}
\caption{Parameters used for the HMC.}
\label{parameters_HMC}
\end{table*}
 
\paragraph{MC dropout.}
We use a dropout probability of $0.05$ and train the predictor network using the log likelihood loss with fixed noise $\sigma_l$ for $2K$ epochs using Adam optimizer with learning rate lr$=10^{-3}$ and weight decay equal to $10^{\frac{-1}{\sqrt{N}}}$ where $N$ is the size of the data.  

\paragraph{Ensemble.} We independently train $10$ models ($5$ for the large UCI datasets) for $3K$ epochs ($500$ for the large UCI datasets) with RMSE loss on mini-batches of size $50$ ($500$ for the large UCI datasets) using stochastic gradient descent (lr$=0.01$, momentum$=0.9$).

\paragraph{MFVI.} 
The parameters of the Gaussian posterior with diagonal covariance matrix were trained using the ELBO as objective and fixed likelihood noise $\sigma_l$.
The ELBO was approximated using $100$ samples for the expected log likelihood and $500$ samples for the Monte Carlo estimate of the KL divergence (using the analytic value of the density of the variational distribution). We used the same mini-batch size, optimizer and learning rate scheme as for NN-HyVI and FuNN-HyVI, except that we had to increase the patience by a factor of two in order to be able to fit the data.

\section{Empirical details and more results}

The values for the fixed noise we used in the likelihood for Exp. 1 (after normalization) are as follows: boston ($\sigma_l=2.5$), concrete ($4.5$), energy ($1.4$), wine ($0.5$), yacht ($1.4$), navalC ($0.2$), powerplant ($3.1$), protein ($4.4$).

To account for variability in training, we independently trained three models for each method except for HMC where we used three random samples of size $1K$ from the 10K samples available.

\textbf{Model evaluation}
For each model considered and every input $X$, we predict a mean value $\bar{y}$ and a probability distribution $p(y|X)$ and variance $\Sigma^2$. For all variational methods, we sampled $1K$ predictors $(f_i)_{i<1K}$ (for HMC, we randomly chose $1K$ predictors among the $10K$ parameters retained). We use the mean and the variance of the sample $(f_i(X))_{i<1K}$ for $\bar{y}$ and $\Sigma^2$, respectively. As for $p(y|X)$, we use $\frac{1}{1000}\sum_{i<1000} p(y|X,f_i)$ where $p(y|X,f_i)$ is the Gaussian likelihood with variance $\sigma_l$ specific to each dataset.

For an Ensemble model $F=(f_i)_{i<N}$ ($N=10$ or $5$) we use similarly the mean and the variance of the sample $(f_i(X))_{i<1K}$ for $\bar{y}$ and $\Sigma^2$, respectively. Moreover we make a Gaussian assumption on $p(y|X)$ and use a normal distribution centered at $\bar{y}$ with variance $\Sigma^2+\sigma_l^2$ $\sigma_l$ is specific to each dataset. 

For MC dropout, we rely on $1K$ forward passes $(y_i)_{i<1K}$ to compute the mean $\bar{y}$ and the variance $\Sigma^2$. As for Ensemble, we use a Gaussian distribution centered at $\bar{y}$ with variance $\Sigma^2+\sigma_l^2$ for $p(y|X)$.

\textbf{Metrics.}
On test data $\mathcal{T}$ the root mean squared error (RMSE) of a model is 
$$\text{RMSE} \sqrt{\tfrac{1}{|\mathcal{T}|}\textstyle\sum_{(X,y)\in \mathcal{T}}(y-\bar{y})^2}\,.$$

The average log posterior predictive (LPP) density of the test data $\mathcal{T}$ (also termed average Log Likelihood) is
$$\text{LPP}=
\tfrac{1}{|\mathcal{T}|}\textstyle\sum_{(X,y)\in \mathcal{T}}\ln p(y|X)\,.
$$

To compute the AUC score associated with the ROC curve we use the NumPy implementation using the predicted epistemic uncertainty $\Sigma^2$ as a score on all available data (in-distribution) versus $10K$ samples from $\nu$ (OOD).

\begin{figure}
\centering
\includegraphics[width=\textwidth]{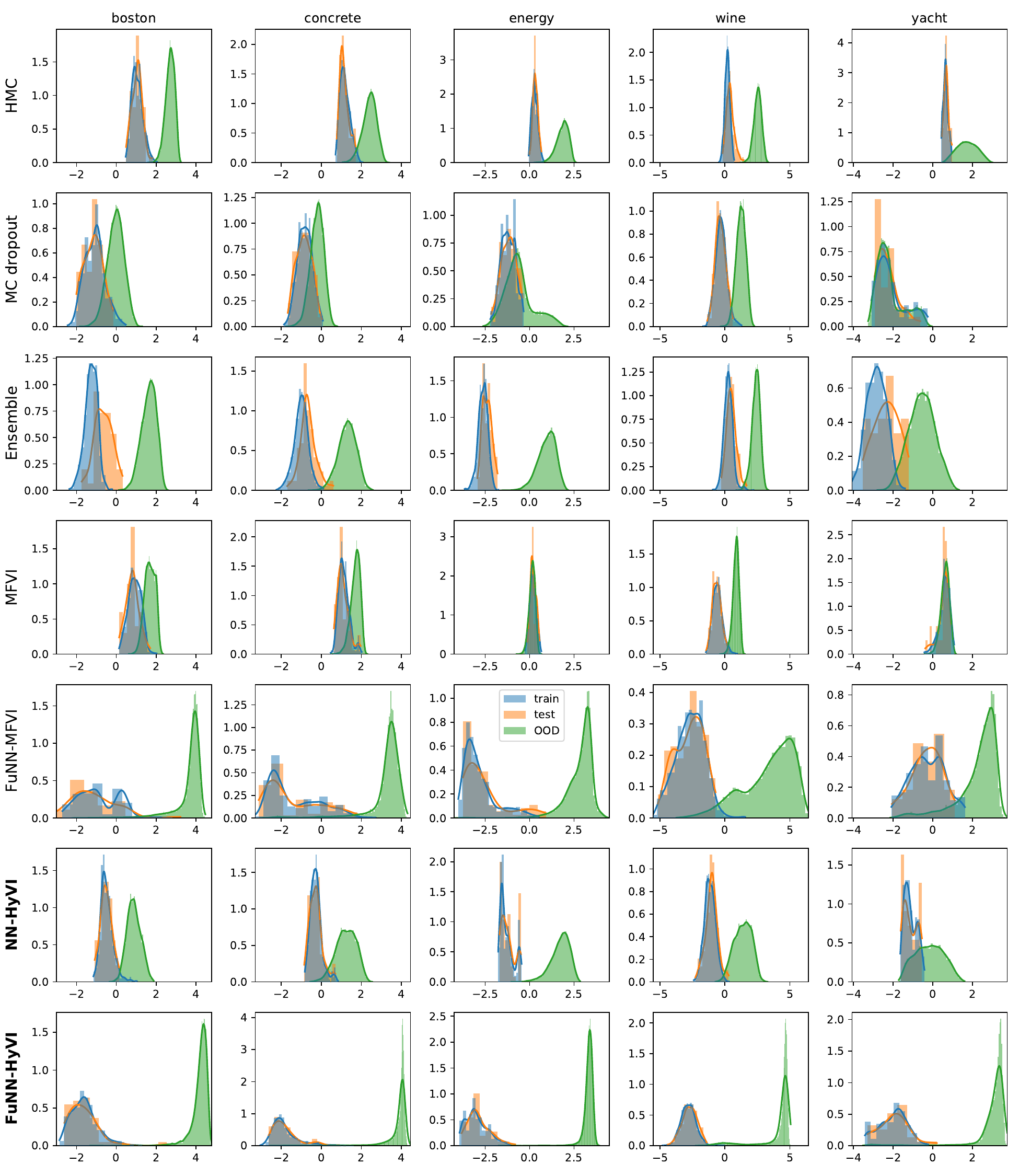}

 \caption{Distributions of epistemic uncertainty on train, test and OOD input for various models on small UCI datasets.}
    \label{fig:SmallUCI_Entropy}

\end{figure}

\begin{figure}[ht]
\centering
\includegraphics[width=\textwidth]{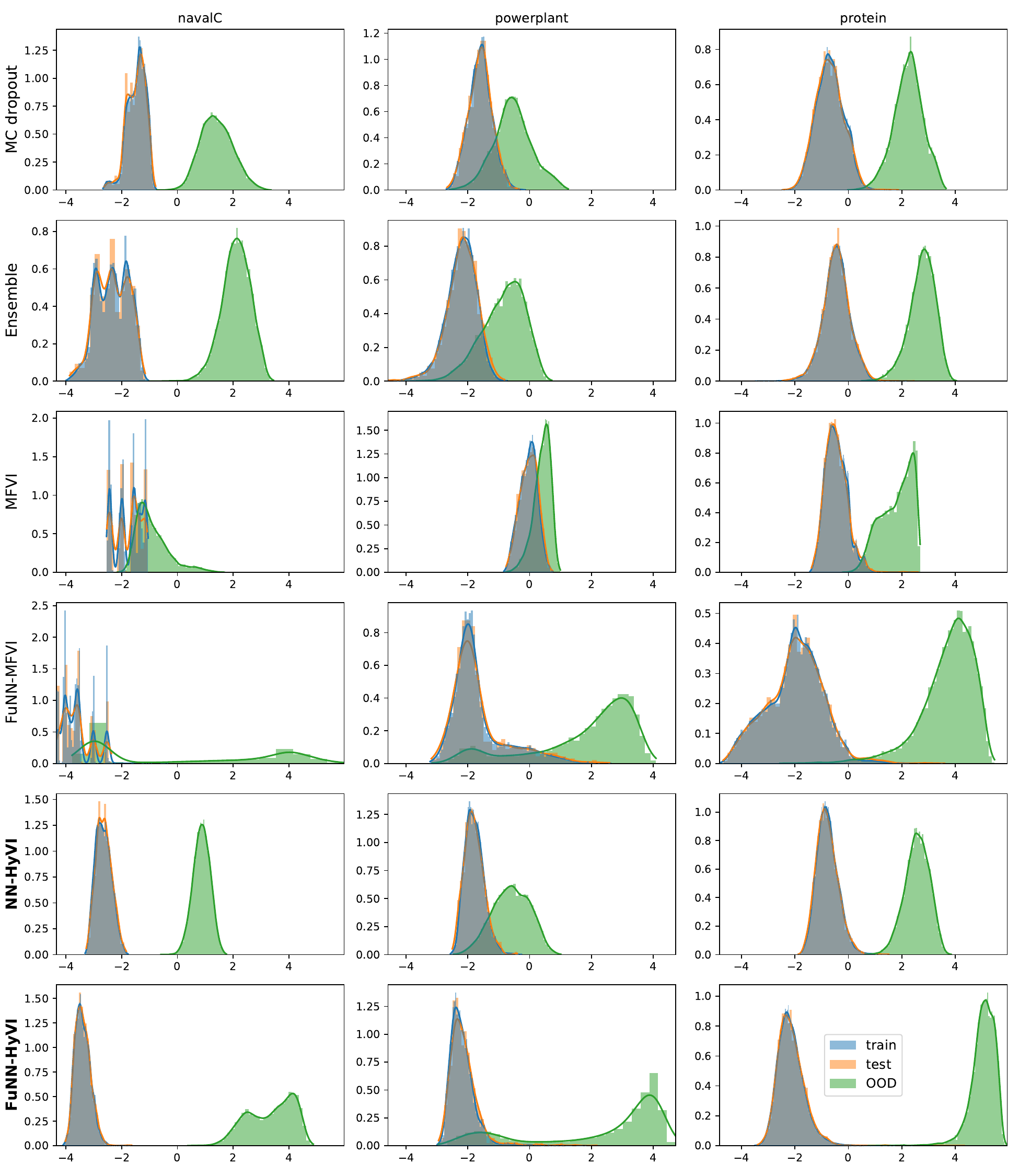}

 \caption{Distribution of epistemic uncertainty on train, test and OOD inputs for large UCI datasets}
    \label{fig:LargeUCI_Entropy}
    
\end{figure}

\textbf{Qualitative results on uncertainty}
The distributions of epistemic uncertainty on train, test and OOD samples are displayed in Fig.~\ref{fig:SmallUCI_Entropy} for small UCI dataset and in Fig.~\ref{fig:LargeUCI_Entropy} for large UCI datasets.

\begin{figure*}[t]
\centering
\includegraphics[width=\textwidth]{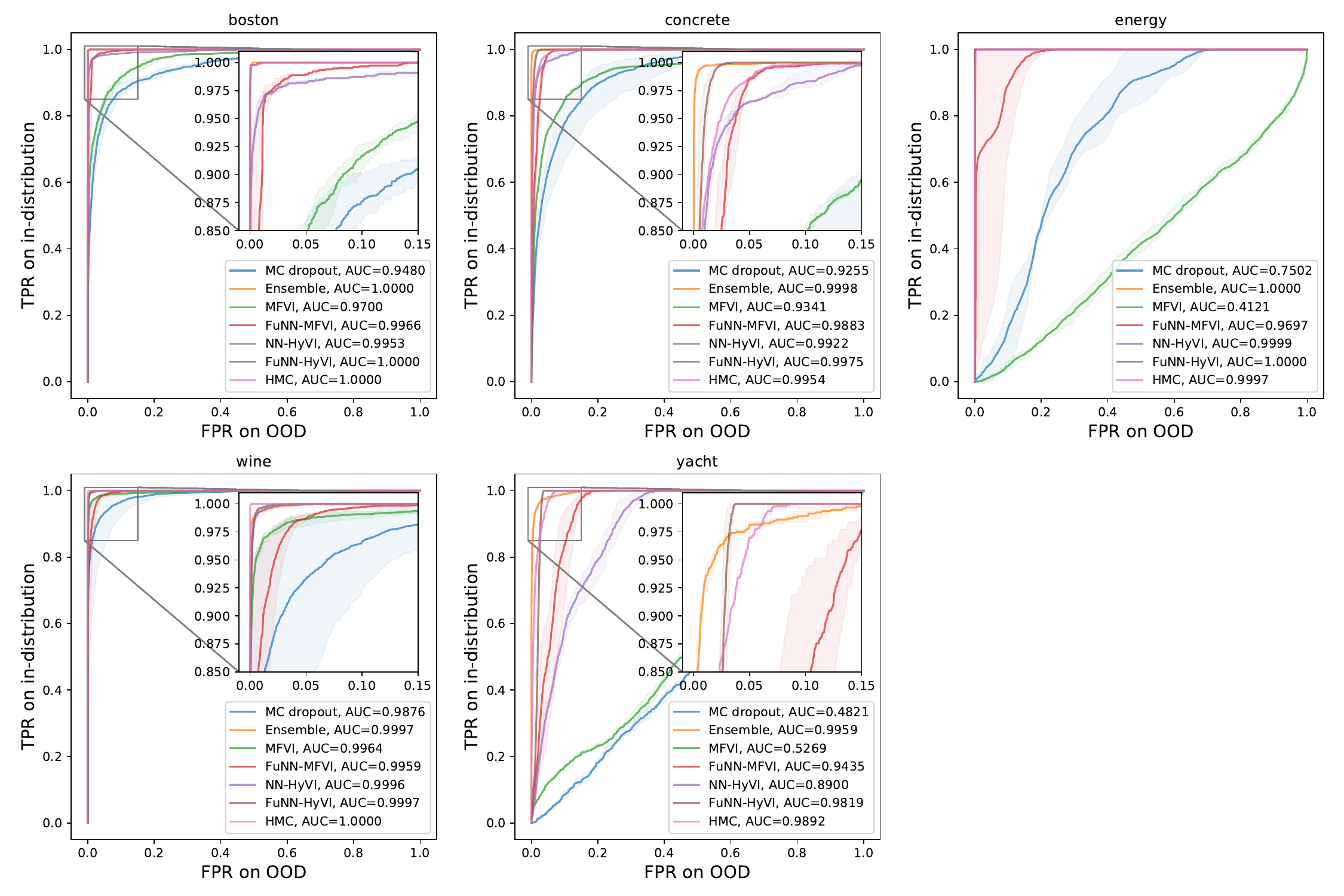}
\caption{ROC curves and AUC for the small UCI datasets.}
    \label{fig:ROC_small}
\end{figure*}

\begin{figure*}[ht]
\centering
\includegraphics[width=\textwidth]{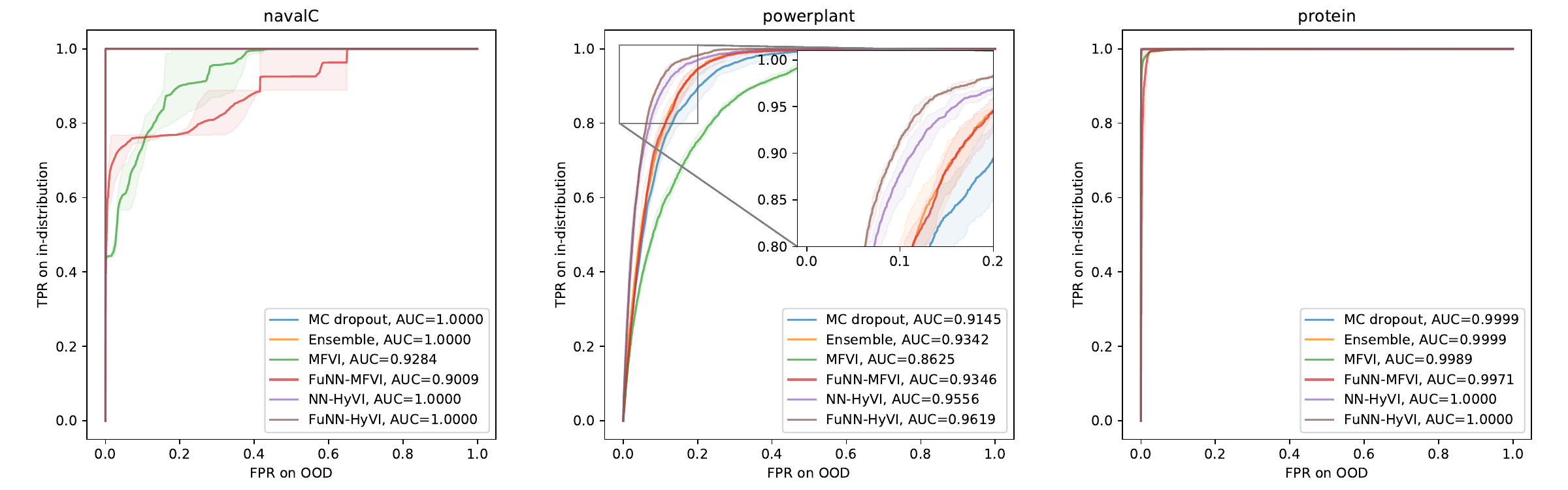}
\caption{ROC curves and AUC for the large UCI datasets.}
    \label{fig:ROC_large}
\end{figure*}

ROC curves are displayed in Fig.~\ref{fig:ROC_small} and Fig.~\ref{fig:ROC_large}. The solid line represents the mean True Positive Rate (TPR) and the shaded area represents the mininimum and maximum TPR over the three models.

We report mean and standard deviation for RMSE and LPP on test data on all UCI datasets in Tables~\ref{tab:RMSELPPsmall} and~\ref{tab:RMSELPPlarge}.

\renewcommand{\std}[1]{\scriptsize$\pm\, \num[round-mode=places,round-precision=2]{#1}$}

\renewcommand{\std}[1]{\scriptsize$\pm\, \num[round-mode=places,round-precision=2]{#1}$}

\begin{table*}[t] 
\center \small
\setlength{\tabcolsep}{3pt}
\begin{tabular}{llllllll}
\toprule
{} &         HMC &  MC dropout &    Ensemble &         MFVI &   FuNN-MFVI &     NN-HyVI &   FuNN-HyVI \\
\midrule
{}& \multicolumn{7}{c}{\bfseries RMSE} \\

\midrule
boston   &   2.63\std{0.02} & \bfseries 2.29\std{0.034} &  2.73\std{0.039} &    3.5\std{0.009} &  4.97\std{0.708} &  2.65\std{0.016} &  3.05\std{0.105} \\
concrete &  6.96\std{0.033} &  5.09\std{0.075} & \bfseries 4.52\std{0.042} &  12.25\std{0.022} &  6.31\std{0.541} &  6.06\std{0.177} &   5.19\std{0.21} \\
energy   &  2.62\std{0.008} &  0.74\std{0.008} & \bfseries 0.41\std{0.003} &   3.41\std{0.005} &  0.79\std{0.141} &  0.68\std{0.014} &  0.49\std{0.009} \\
wine     &    0.58\std{0.0007} &   0.60\std{0.009} & \bfseries 0.57\std{0.004} &    0.62\std{0.01} &   0.7\std{0.002} &  0.73\std{0.006} &   0.72\std{0.03} \\
yacht    &  3.84\std{0.022} &  1.06\std{0.043} & \bfseries 0.86\std{0.035} &   3.89\std{0.028} &  2.25\std{0.184} &  1.28\std0.044 &  1.07\std{0.089} \\

\midrule

{}& \multicolumn{7}{c}{\bfseries LPP} \\
\midrule
boston   &  -4.1\std{2e-04} &  -4.1\std{0.00014} & \bfseries  -3.2\std{0.2008} &  -4.1\std{2e-04} &  -4.1\std{0.01011} &  -4.1\std{2e-04} &  -4.1\std{9e-4} \\
concrete &   -5.3\std{1e-04} &  -5.2\std{7e-05} &  \bfseries    -3.6\std{0.1590} &  -5.3\std{2e-04} &  -5.2\std{8e-04} &  -5.2\std{2e-04} &  -5.2\std{2e-04} \\
energy   &   -3.6\std{2e-04} & -3.6\std{1e-04} &  \bfseries -1.6\std{0.47242} &  -3.6\std{2e-04} &  -3.6\std{6e-04} &  -3.6\std{2e-04} &    -3.6\std{2e-05} \\
wine     & \bfseries -0.9\std{7e-04} &  -1.0\std{0.02606} &  -1.5\std{0.06858} &  -1.1\std{0.03673} &  -1.5\std{0.00982} &   -1.5\std{0.0522} &  -1.6\std0.13075 \\
yacht    &  -4.1\std{5e-04} &   -4.0\std{1e-04} & \bfseries -0.5\std{0.08407} &  -4.1\std{6e-04} &  -4.0\std{3e-03} &    -4.0\std{7e-05} &    -4.0\std{8e-05} \\

\bottomrule
\end{tabular}
    \caption{Average and standard error for RMSE and LPP for small UCI datasets.
    }
    \label{tab:RMSELPPsmall}
\end{table*}

\begin{table*}[t] 
\center \small
\setlength{\tabcolsep}{2pt}
\begin{tabular}{lllllll}
\toprule
{} &           MC dropout &    Ensemble &         MFVI &   FuNN-MFVI &     NN-HyVI &   FuNN-HyVI \\
\midrule
{}& \multicolumn{6}{c}{\bfseries RMSE} \\
\midrule
navalC     &    0.00038\std{1e-05} &    0.00041\std{2e-05} &     0.0005\std{5e-05} &  0.00095\std{2e-04} &   \bfseries 0.00025\std{3e-05} &    0.00025\std{3e-05} \\
powerplant & \bfseries 3.69367\std{0.01355} &  3.69755\std{3e-03} &  4.04913\std{0.02809} &   4.17883\std{0.0754} &  3.87214\std{0.01769} &  3.71361\std{0.01161} \\
protein    & 4.38351\std{9e-04} & \bfseries  4.1685\std{0.01109} &  4.88476\std{0.00709} &  4.48941\std{0.06054} &   4.5195\std{0.01208} &  4.25881\std{0.01487} \\
\midrule
{}&\multicolumn{6}{c}{\bfseries LPP} \\

\midrule
navalC     &   6.07\std{0.02685} &   \bfseries   6.26\std{0.01591} &  -8.27\std{3.0823} &  -42.46\std{16.27946} &  4.71\std{0.72598} &  4.24\std{1.16301} \\
powerplant &   \bfseries -4.89\std{1e-05} &  -60.64\std{10.19581} &   \bfseries -4.89\std{4e-05} &    \bfseries -4.89\std{4e-04} &  \bfseries   -4.89\std{2e-05} &  \bfseries  -4.89\std{2e-05} \\
protein    &   -4.23\std{1e-05} &   -16.72\std{0.81962} &   -4.23\std{6e-05} &    -4.23\std{7e-04} &  -4.23\std{1e-04} &   \bfseries -4.22\std{9e-05} \\
\bottomrule
\end{tabular}
    \caption{Average and standard error for RMSE and LPP for large UCI datasets.
    }
    \label{tab:RMSELPPlarge}
\end{table*}

\section{Comparison of running time}

Table~\ref{tab:runningtime} shows average running time in seconds on GPU (NVIDIA GeForce RTX 2080 Ti).

\begin{table}[ht!]
\vspace{-2mm}
\small
\centering

\begin{tabular}{lccccccc}
\toprule
{} & $\text{HMC}^{\star}$  & MC dropout &  Ensemble &   MFVI &  FuNN-MFVI &  NN-HyVI &  FuNN-HyVI \\
\midrule
boston   &  30K &        67 &     193 &  118 &      238 &    138 &      246 \\
concrete &  45K  &      123 &     378 &  169 &      351 &    264 &      405 \\
energy   &   30K &       99 &     271 &  192 &      349 &    220 &      331 \\
wine     &   80K &    183 &     459 &  569 &      573 &    642 &      533 \\
yacht    &   25K  &      43 &     132 &  105 &      138 &    108 &      184 \\
\bottomrule
\end{tabular}

    \caption{Average running time  on GPU in seconds. \\
    $^\star:$ We also included running times for HMC for the sake of comparison, but note that they were executed on CPUs and no real effort has been made to optimize the running time.}
    \label{tab:runningtime}
\end{table}

\end{document}